\documentclass[12pt,journal,compsoc,onecolumn]{IEEEtran}
\usepackage{amsmath,graphicx}
\usepackage{amssymb,amsfonts}
\usepackage{dblfloatfix}
\usepackage{multirow}
\usepackage[noadjust]{cite}

\usepackage[ruled,vlined]{algorithm2e}
\usepackage{graphicx}
\newcommand{\cmark}{\ding{51}}
\newcommand{\xmark}{\ding{55}}
\usepackage{pifont}

\usepackage{booktabs}
\usepackage{siunitx}
\usepackage[numbers,compress]{natbib}
\usepackage{texnames}
\usepackage{bm,bbm}

\usepackage{lipsum}

\def\x{{\mathbf x}}

\def\XX{{\bf X}}

\def\x{{\bf x}}

\def\D{{d}}

\def\F{{ f}}
\def\DD{{d}}
\def\tr{{\bf tr}}
\def\1{{\bf 1}}

\def\x{{\bf x}}  
  
\def\V{{\bf D}}
\def\VV{{\cal V}}

\def\G{{\cal G}}
\def\W{{\bf W}}
\def\III{{\bf I}}
\usepackage{amsthm}
\newtheorem{proposition}{Proposition}
\newtheorem{definition}{Definition}

\usepackage[ruled,vlined]{algorithm2e}

\newlength{\bibitemsep}
\newlength{\bibparskip}
\let\oldthebibliography\thebibliography
\renewcommand\thebibliography[1]{
  \oldthebibliography{#1}
  \setlength{\parskip}{\bibitemsep}
  \setlength{\itemsep}{\bibparskip}
}

\title{Label-Efficient Skeleton-based Recognition with Stable-Invertible Graph Convolutional Networks}
\author{Hichem Sahbi \\
$ $ \\
Sorbonne University, CNRS, LIP6,  F-75005, Paris, France 
 }

\begin{document}
 \maketitle
\begin{abstract}
Skeleton-based action recognition is a hotspot in image processing. A key challenge of this task lies in its dependence on large, manually labeled datasets whose acquisition is costly and time-consuming. This paper devises a novel, label-efficient method for skeleton-based action recognition using graph convolutional networks (GCNs). The contribution of the proposed method resides in learning a novel acquisition function --- scoring  the most informative subsets for labeling --- as the optimum of an objective function mixing data representativity, diversity and uncertainty. We also extend this approach by learning the most informative subsets using an invertible GCN which allows mapping data from ambient to latent spaces where the inherent distribution of the data is more easily captured. Extensive experiments, conducted on two challenging skeleton-based recognition datasets, show the effectiveness and the outperformance of our label-frugal GCNs against the related work.
\end{abstract}

\section{Introduction}
\label{sec:intro}
Skeleton-based recognition is a major task in image processing which consists in analyzing skeletal structures (human body, hands, etc.) by extracting joint positions and modeling their interactions. This task is particularly useful in challenging scenarios like cluttered environments. 
Early methods rely on handcrafted features~\cite{Yun2012,refref18,Ji2014,Li2015a,refref40,refref41,refref59,refref39}, such as joint angles and relative distances, fed into classifiers like support vector machines and hidden Markov models \cite{Vemulapalli2014,refref11,Sahbiling2013,Sahbiyuan2012,Sahbimazari2019} as well as manifold learning~\cite{RiemannianManifoldTraject,refref61,Huangcc2017,ref23}. With the rise of deep learning \cite{refref10}, recurrent neural networks (RNNs), particularly LSTMs and GRUs \cite{Du2015,Liu2016,Zhua2016,Zhang2017b,GCALSTM,DeepGRU}, became popular for capturing the temporal dynamics of skeletal sequences. Graph Convolutional Networks (GCNs) have also emerged as powerful learning models, exploiting the inherent graph structures of skeletons to learn spatial relationships between joints, as a part of attention-based models \cite{Jiang2020,Song2017,Liu2021,refffabc888,sahbi2021iccv}. The latter have been demonstrating impressive performances by effectively modeling long-range dependencies and capturing complex motion patterns.\\

\indent The success of the aforementioned learning-based methods, for skeleton-based recognition, hinges on the availability of large, diverse datasets of hand-labeled skeleton sequences. However, acquiring such massive data sequences is known to be time and labor demanding. Several solutions address this issue including data augmentation \cite{DA2}, few shot and transfer learning \cite{TL1}, as well as self-supervised learning \cite{SS1}. Nonetheless, the relative success of these solutions relies upon a strong assumption that knowledge are enough in order to close the {\it accuracy gap} while actually labeled data are  more important. Another category of methods is active learning (AL) \cite{Burr2009,sahbi2023aactive} which excels at adapting to the ``oracle'' (expert annotator) in a way that other label-efficient methods do not. Indeed, unlike other methods, AL queries only the {\it most informative} unlabeled samples for annotation by quantifying and maximizing the impact of labeling a particular sample on a learning model. Informative datasets are usually selected based on various criteria, notably diversity \cite{Ref50a,Ref51a} and uncertainty \cite{SS1,Ref54,Ref56,Ref57,Culotta2005} in different contexts \cite{du2022efficient,cai2022self,kondo2023active}. Uncertainty-based methods include margin sampling and entropy-based criteria \cite{zhao2023uncertainty,kim2023batch} while diversity-based approaches include coverage maximization \cite{wu2022adaptive,zhang2022multi}.  Other strategies consider the representativeness of data, selecting samples that are most similar to the overall data distribution. However, current approaches for identifying these informative subsets often rely on heuristics, lacking rigorous theoretically grounded framework. This limits the optimality of the selected data and can hinder the overall efficiency of AL.\\

\indent Considering the aforementioned issues, we introduce in this paper a label-efficient GCN for skeleton-based recognition. The contribution of the proposed method resides in a novel principled probabilistic framework that designs unlabeled exemplars (candidate samples for labeling) instead of sampling them from a fixed pool of unlabeled data. These exemplars are obtained as an interpretable solution of an objective function mixing data representativity, diversity and uncertainty. Our proposed framework designs these exemplars using a stable and an invertible GCN that allows mapping input graphs (lying on highly nonlinear manifolds) from ambient (input) to latent spaces where designing these exemplars becomes more tractable; indeed, with the proposed GCNs, data in the latent space follow a standard probability distribution (namely gaussian) whose sampling and search is more tractable compared to the arbitrary distributions in the ambient space. Once designed, the learned exemplars are mapped back to the input space thanks to the invertibility and stability of our designed GCNs. Extensive experiments, conducted on two challenging skeleton-based recognition tasks,  show the outperformance of our label-efficient  method compared to the related work.

\def\UU{{\cal U}}
\def\DD{{\cal D}}
\def\YY{{\cal Y}}
\def\HH{{\bf H}}

\section{Proposed  Display Model}
Our proposed AL solution consists of two principal building blocks: {\it display} and {\it learning} models. The {\it former} aims at designing an acquisition function probing an oracle about the labels of the most informative data, whilst the {\it latter} seeks to retrain a label-frugal classifier accordingly. These two steps are iteratively applied till reaching enough classification accuracy or exhausting a predefined labeling budget.  Let   $\UU=\{\x_1,\dots,\x_n\} \subset \mathbb{R}^p$ be a pool of unlabeled data; at each AL iteration $t \in \{0,\dots,T-1\}$, a subset $\DD_t$ --- referred to as {\it display} --- is built from $\UU$ (following the model in section~\ref{display}), and used to query the oracle about its labels $\YY_t$.  Then a classifier $f_t$ is trained on  $\cup_{k=0}^t (\DD_k,\YY_k)$.

Our first contribution (introduced in section~\ref{display}) is based on a novel display model that builds in a {\it flexible way} displays instead of sampling fixed ones from $\UU$.  

\subsection{Display model design}\label{display} 
The principle of our method consists in designing the most diverse, representative and uncertain data that challenge (the most) the current classifier $f_t$, leading to a better re-estimate of $f_{t+1}$ in the subsequent iteration $t+1$ of active learning. We consider a probabilistic framework that {\it builds} the subsequent display $\DD_{t+1}$ (denoted for short as $\DD$) instead of sampling $\DD$ from $\UU$. Let $\XX \in \mathbb{R}^{p \times n}$ (resp. $\V \in \mathbb{R}^{p \times K}$) be a matrix whose k-th column $\XX_k$ (resp. $\V_k$) corresponds to an element of $\UU$ (resp. $\DD$) and $K=|\DD|$. In order to obtain the display $\V$, our proposed framework assigns for each $\V_k$, a conditional probability distribution measuring the memberships $\{\mu_{ik}\}_i$ as the contribution of each $\x_i \in \UU$ in shaping $\V_k$. These memberships $\mu=\{\mu_{ik}\}_{ik}$ and the display $\V$ are found by minimizing the following constrained objective function
\def\tr{{\bf tr}}
\begin{equation}\label{of} 
  \begin{array}{cc}
\displaystyle  \min_{\mu \in \Omega,\V} &  \displaystyle \tr(\mu \ \D(\XX,\V)^\top)  + \alpha \ \sum_{k,k'}^{K,N}  \exp\big(-\frac{1}{\sigma}\big\| \V_k - \HH_{k'}\big\|_2^2\big)  \\
           & + \ \beta \ \tr(\V^\top \V)  + \gamma \  \tr(\mu^\top \log \mu),
\end{array}   
\end{equation} 
being  $\Omega=\{\mu :  \mu  \geq 0; \1_n^\top \mu  = \1_K^\top\}$ a convex set that constrains $\mu$ to be column-stochastic (i.e., each column as a conditional probability distribution), $^\top$ denotes the transpose, and $\1_{K}$, $\1_{n}$ are two vectors of $K$ and $n$ ones respectively. The first term of Eq.~\ref{of} encodes the representativeness of the designed exemplars in $\V$, aiming to minimize the discrepancy between these exemplars  and the original distribution of data in $\UU$. It also serves to constrain the oracle's annotations only on realistically designed exemplars, thereby ensuring relevant annotations and also preventing the selection of trivial or meaningless exemplars. The second term of Eq.~\ref{of} captures diversity of $\V$; this term seeks to maximize the difference between the $N$ previously and the $K$ currently designed exemplars (resp. matrices $\HH$ and $\V$), and enforces the new ones to be as far as possible from the previous ones.

The third term of  Eq.~\ref{of} acts as an equilibrium criterion measuring the uncertainty associated with exemplars in $\V$; in other words, it encourages 
 exemplars to lie on the decision boundaries of the learned classifiers, and it also acts as a regularizer on $\V$. Minimizing this term effectively identifies exemplars which are inherently ambiguous, and targeting annotations on these highly uncertain data is crucial to reduce model ambiguity and to speedup convergence to  well-defined decision functions. Finally, the fourth term corresponds to a regularizer on $\mu$ which  considers that without any a priori on the three other terms, the conditional probabilities $\mu=\{\mu_{ik}\}_{ik}$  should be flat. All the aforementioned terms are weighted by $\alpha$, $\beta$, $\gamma \geq 0$ whose setting is described subsequently.

\def\H{{\bf H}}
\begin{proposition}
The optimality conditions of Eq.~\ref{of} leads to the solution as the fixed-point of  
\begin{equation}\label{eq2}
\begin{array}{lll}
  \mu^{(\tau+1)}& :=&\displaystyle   \hat{\mu}^{(\tau+1)} \  \textrm{\bf diag} \big( \1_n^\top \hat{\mu}^{(\tau+1)}\big)^{-1} \\
  \V^{(\tau+1)} &:= & \hat{\V}^{(\tau+1)} \  \big(\textrm{\bf diag} (\1^\top_n {\mu}^{(\tau)}) + \beta \III \big)^{-1},

\end{array}  
\end{equation}
being  $\hat{\mu}^{(\tau+1)}$, $\hat{\V}^{(\tau+1)}$ respectively
\begin{equation}\label{eq3} 
\begin{array}{l}
  \exp\big\{-\frac{1}{\gamma}d(\XX,\V^{(\tau)})\big\},\\
  \\
  \displaystyle \XX \ \mu^{(\tau)}  - \frac{2\alpha}{\sigma}  \big( \V^{(\tau)} \  \textrm{\bf diag}(\1_N' {\bf S}) - \H {\bf S}\big),
\end{array} 
\end{equation}
where  ${\bf S}$ equates (with $\V^{(\tau)}$ written for short as $\V$)
\begin{equation}\label{eqqqabc}
\displaystyle  \exp\bigg\{-\frac{1}{\sigma} \big(\1_N \textrm{\bf diag}(\V^\top\V)^\top+\textrm{\bf diag}(\H^\top\H) \1_K^\top-2\H^\top\V\big)\bigg\},
\end{equation}
here ${\bf S}$ is a similarity matrix between $\V$ and $\HH$, $\1_N$ is a vector of $N$ ones, and $\textrm{\bf diag}$ maps a vector to a diagonal matrix. 
\end{proposition} 
In view of space, details of the proof are omitted and follow the optimality conditions of Eq.~\ref{of}'s gradient. More importantly, the solution of $\mu$ in Eq.~\ref{eq3} shows that low distances lead to high memberships of the input data in $\XX$ to the underlying exemplars in $\V$, and vice versa, whereas the solution of $\V$ shows that each exemplar $\V_k$ is defined as a combination of two terms: the first one as a normalized\footnote{thanks to the column-stochasticity of $\mu$.} linear combination of actual data weighted by their memberships to $\V_k$ whilst the second term disrupts further $\V_k$ to make it as different as possible from the previously designed exemplars in $\HH$ (depending on the setting of $\alpha$). 
\noindent Note that ${\mu}^{(0)}$ and  ${\V}^{(0)}$ are initially set to random values and, in practice, the procedure converges to an optimal solution (denoted as $\tilde{\mu}$, $\tilde{\V}$) in few iterations. This solution defines the subsequent display $\DD_{t+1}$ used to train $f_{t+1}$. Note also that $\alpha$ and $\beta$ are set to make the impact of the underlying terms equally proportional, and this corresponds to $\alpha=\frac{1}{KN}$ and $\beta=\frac{1}{Kp}$. In Eq.~\ref{eq3}, the hyperparameter $\sigma$ is set proportionally to $\alpha$ in order to absorb the former by the latter, and thereby reduce the total number of hyperparameters. Finally,  since $\gamma$ acts as  scaling factor that controls the shape of the exponential function, its setting is iteration-dependent and  proportional to the input of that  exponential (i.e.,  $\log(\hat{\mu}^{(\tau+1)})$),  so in practice  $\gamma= \frac{1}{nK} \|\log(\hat{\mu}^{(\tau+1)})\|_1$. \\
\textcolor{black}{Now considering the foregoing  AL formulation, two variants of the proposed solution are considered in this paper. The first one finds exemplars using the above formulation directly in the ambient (input) space, while the second one finds the exemplars in the latent space, and maps them back to the ambient space thanks to the invertibility and also stability of the learned GCNs (as shown in section~\ref{learning}). As shown subsequently, relying on invertible and stable GCN mapping leads to an extra gain in AL performances as also shown later through experiments}.

\def\SS{{\cal S}}
\def\E{{\cal E}}
\def\A{{\bf A}}
\def\F{{\cal F}}
\def\UU{{\bf U}} 
\def\W{{\bf W}}
\def\WW{{\bf W}}
\def\N{{\cal N}}
\section{Proposed Learning Model}\label{learning}
As introduced, the success of the aforementioned active learning process is highly reliant on the suitability of the display model. In other words, finding suitable displays in the input space should reflect the distribution of the data in the input space. However, for arbitrary input data distributions the display model, in Eq.~\ref{of}, may hit a major limitation; input data lying on nonlinear manifolds are challenging to parse in order to guarantee that designed displays still lie on these manifolds. In the sequel of this section, we revisit GCNs and we introduce  --- as a second contribution --- a novel design that makes our trained GCNs invertible and stable.
\subsection{A Glimpse on graph convnets} Consider a collection of graphs $\{\G_i=(\VV_i, \E_i)\}_i$, where $\VV_i$ and $\E_i$ represent the nodes and edges $\G_i$, respectively.  For simplicity, let $\G=(\VV, \E)$ denote a single graph from this collection.  Each graph $\G$ is associated with a signal $\{\psi(v) \in \mathbb{R}^s: \ v \in \VV\}$ and an adjacency matrix $\A$. Graph Convolutional Networks (GCNs) aim to learn a set of $C$ filters $\F$ that define a convolution operation on the $m$ nodes of $\G$ (where $m=|\VV|$) as follows: $(\G \star \F)_\VV = g\big(\A \  \UU^\top  \   \W\big)$. Here, $\UU \in \mathbb{R}^{s\times m}$ is the graph signal, $\W \in \mathbb{R}^{s \times C}$ is the matrix of convolutional parameters for the $C$ filters, and $g(.)$ is a nonlinear activation function applied element-wise.  In this operation, the input signal $\UU$ is projected using the adjacency matrix $\A$, effectively aggregating the signals from the neighbors of each node $v$. The entries of $\A$ can be either handcrafted or learned. Hence, $(\G \star \F)_\VV$ can be viewed as a two-layer (attention and convolutional) block. The first layer aggregates signals from the neighborhood $\N(\VV)$ of each node by multiplying $\UU$ with $\A$, while the second layer performs the convolution by multiplying the resulting aggregates with the $C$ filters in $\W$. 
\subsection{Invertibility \& Stability}
In what follows, we formally subsume a given GCN as a multi-layered neural network $f$  whose weights are defined as $\theta =  \left\{\WW_1,\dots, \WW_L \right\}$, being $L$ its depth,  $\WW_\ell \in \mathbb{R}^{d_{\ell-1} \times d_{\ell}}$ its $\ell^\textrm{th}$  layer weight tensor, and $d_\ell$ the dimension of $\ell$. The output of a given layer  $\ell$ is defined as
$ \mathbf{\phi}^{\ell} = g_\ell(\WW_\ell^\top \  \mathbf{\phi}^{\ell-1})$, $\ell \in \{2,\dots,L\}$,  with $g_\ell$ an activation function; without a loss of generality, we omit the bias in the definition of  $\mathbf{\phi}^{\ell}$.\\
In this section, we are interested in designing invertible and stable networks. Invertibility (bijection) of  $f: \mathbb{R}^p \rightarrow \mathbb{R}^q$ guarantees the existence of  a {\it one-to-one} mapping from  $\mathbb{R}^p$ to  $\mathbb{R}^q$ (with necessarily $p=q$)\footnote{As the output of $f$ depends on the number of classes, a simple trick consists in adding fictitious outputs to match any targeted dimension (similarly for other layers).}  so as no distinct network's inputs $\phi_1^1$,  $\phi_2^1$ map to the same output $\phi^L$,  and  for every output  $\phi^L$, there exists at least one  input   $\phi^1$ such that $f(\phi^1) = \phi^L$. Stability pushes invertibility ``one step further'' to guarantee that $f^{-1}$ --- when evaluated on a given targeted latent distribution (e.g., gaussian) --- does not diverge from the ambient (input) distribution.  

\begin{definition}[Stability] An invertible  network  $f : \mathbb{R}^p  \rightarrow \mathbb{R}^q$ is called bi-Lipschitzian (or KM-Lipschitzian), if $f$ is K-Lipschitzian and its inverse  $f^{-1}$ is  M-Lipschitzian. 
\end{definition}

\noindent In general, making both $K$ and $M$ small for any given nonlinear function is challenging \cite{heinonen2005lectures}. However, considering our following network $f$'s design, it becomes possible under specific conditions to make both $K$ and $M$ small (namely close to 1 as a result of our subsequent proposition).

\begin{proposition}
Provided that  (i) the entrywise  activations  $\{g_\ell(.)\}_{\ell=2}^L$  are bijective in $\mathbb{R}^p$,   (ii) $l \leq |g'_\ell(.)|\leq u$,    and (iii) all the weight matrices in  $\theta$ orthonormal,  then the  network  $f$ is invertible in $\mathbb{R}^p$,  and KM-Lipschitzian with $K=u^{L-1}$ and $M=(1/l)^{L-1}$. 

\end{proposition} 

\noindent  Details of the proof are given in the appendix.   More importantly, following the above proposition, when $f$ is invertible in $\mathbb{R}^p$, then one may derive $f^{-1}(\phi^{L})=\phi^{1}$ being  $\phi^{\ell-1}=(\W_\ell^\top)^{-1} g_\ell^{-1}(\phi^{\ell})$,  and when $l$ and $u$ are close to $1$,  then  $K,  M \approx 1$ meaning that both $f$ and  $f^{-1}$ are  $1$-Lipschitzian \cite{heinonen2005lectures} so any slight update of  exemplars in the latent space (with the fixed-point iteration in Eq.~\ref{eq2}) will also result into a slight update of these exemplars in the ambient space when applying $f^{-1}$. This eventually leads to stable  exemplar design in the ambient space, i.e., they follow the actual distribution of data manifold. As a Lipschitz constant of $f$ is $\prod_{\ell} \|\W_\ell \|_2 . \big|g'_\ell\big|$,  and for $f^{-1}$ is $\prod_{\ell} \|(\W_\ell^\top)^{-1}\|_2 \ |{g^{-1}_\ell}'|$  (see proof in appendix), the sufficient conditions that guarantee that both $f$ and $f^{-1}$ are Lipschitzian (with $K,  M\approx 1$) corresponds to (1) $\|\W_\ell \|_2 \approx 1$, and (2) $l, u \approx 1$ with $l<u$. Hence, by design, conditions (1)+(2) could be satisfied by choosing the slope of the activation functions to be close to one (in practice $u=0.99$ and $l=0.95$ corresponding  respectively to  the positive and negative slopes  of the leaky-ReLU), and also by constraining all the weight matrices to be {\it orthonormal} which also guarantees their invertibility. This is obtained by adding a regularization term, to the cross-entropy (CE) loss, when training GCNs, as

\begin{equation}\label{eqqqqq} 
\min_{\{\W_\ell\}_\ell}{\textrm{CE}}(f;\{\W_\ell\}_\ell) + \lambda \ \sum_{\ell} \big\|\W_\ell^\top \W_\ell-\III\big\|_F,
\end{equation}
here $\III$ stands for identity, $\|.\|_F$ denotes the Frobenius norm and $\lambda>0$ (with $\lambda=\frac{1}{p}$ in practice\footnote{Note that at frugal data regimes, this optimization problem is easy to minimize as the cross entropy term involves few labeled data, so it is enough to set $\lambda$ to small values in order to guarantee the minimization of both terms.}); in particular, when $\W_\ell^\top \W_\ell-\III =0$,  then $\W_\ell^{-1}=\W_\ell^\top$ and  $\|\W_\ell \|_2 =\|\W_\ell^{-1} \|_2=1$.  With this formulation, the learned GCNs are guaranteed to be discriminative, invertible and stable.

\section{Experiments}\label{sec:experiments}
This section evaluates the performance of baseline and label-frugal GCNs for skeleton-based recognition using the SBU Interaction \cite{Yun2012} and First Person Hand Action (FPHA) \cite{refref11}  datasets. The SBU Interaction dataset, captured using the Microsoft Kinect, comprises 282 skeleton sequences of two interacting individuals performing one of eight predefined actions. Each interaction is represented by two 15-joint skeletons, with each joint's 3D coordinates acquired across the video frames.  Evaluation follows the original train-test split defined in \cite{Yun2012}. The FPHA dataset contains 1175 skeleton sequences spanning 45 diverse hand-action categories, performed by six individuals across three different scenarios.  The actions exhibit significant variations in style, speed, scale, and viewpoint. Each skeleton consists of 21 hand joints, also represented by sequences of 3D coordinates. Following \cite{refref11}, we evaluate performance using the 1:1 setting, with 600 sequences for training and 575 for testing. For both datasets, we report the average classification accuracy across all action categories. \\
\noindent {\bf Input graphs.} We represent each skeleton sequence  $\{S^t\}_t$ as a series of 3D joint coordinates $S^t=\{\hat{p}_j^t\}_j$ at each frame $t$. A joint's trajectory $\{\hat{p}_j^t\}_t$ tracks its movement across frames.  Our input graph $\G = (\VV,\E)$ comprises nodes $\VV$ with each one $v_j \in \VV$ representing a trajectory $\{\hat{p}_j^t\}_t$, and each edge  $(v_j, v_i) \in  \E$ connects spatially neighboring trajectories.  To process each trajectory, we divide its duration into $M_c$  equal temporal chunks (with $M_c=4$ in practice). Joint coordinates $\{\hat{p}_j^t\}_t$ are assigned to these chunks based on their timestamps, and the average coordinates within each chunk are concatenated to form a trajectory descriptor (denoted as $\psi(v_j) \in \mathbb{R}^{s}$) of size  $s=3M_c$. This chunking approach preserves temporal information while making the representation independent of frame rate and sequence duration. \\
\begin{table}  
  \begin{minipage}[c]{0.85\columnwidth}
\centering
 \begin{center}
\resizebox{0.62\columnwidth}{!}
{
\begin{tabular}{cc|c}
{\bf Method}      &   & {\bf Accuracy (\%)}\\
\hline 
  Raw Position \cite{Yun2012} & $ \ $   & 49.7   \\ 
  Joint feature \cite{Ji2014}  & $ \ $   & 86.9   \\
  CHARM \cite{Li2015a}       & $ \ $    & 86.9   \\
 \hline  
H-RNN \cite{Du2015}         & $ \ $    & 80.4   \\ 
ST-LSTM \cite{Liu2016}      & $ \ $    & 88.6    \\ 
Co-occurrence-LSTM \cite{Zhua2016} & $ \ $  & 90.4  \\ 
STA-LSTM  \cite{Song2017}     & $ \ $   & 91.5  \\ 
ST-LSTM + Trust Gate \cite{Liu2016} & $ \ $  & 93.3 \\
VA-LSTM \cite{Zhang2017}      & $ \ $  & 97.6  \\
 GCA-LSTM \cite{GCALSTM}                    &   $ \ $      &  94.9     \\ 
  \hline
Riemannian manifold. traj~\cite{RiemannianManifoldTraject} &  $ \ $  & 93.7 \\
DeepGRU  \cite{DeepGRU}        &    $ \ $   &    95.7    \\
RHCN + ACSC + STUFE \cite{Jiang2020} & $ \ $   & 98.7 \\ 
  \hline
\hline 
  \textcolor{black}{Our baseline GCN} &              &        98.4      
\end{tabular}}
 \end{center} 
\caption{Comparison of our baseline GCN (not label-efficient) against related work on the SBU database.}\label{tab222}
\vspace{0.25cm}
\end{minipage}
\begin{minipage}[c]{0.8\columnwidth}
  \centering
 \begin{center}
\resizebox{0.79\columnwidth}{!}{
\begin{tabular}{ccccc}
{\bf Method} & {\bf Color} & {\bf Depth} & {\bf Pose} & { \bf Accuracy (\%)}\\
\hline
  2-stream-color \cite{refref10}   & \cmark  &  \xmark  & \xmark  &  61.56 \\
 2-stream-flow \cite{refref10}     & \cmark  &  \xmark  & \xmark  &  69.91 \\  
 2-stream-all \cite{refref10}      & \cmark  & \xmark   & \xmark  &  75.30 \\
\hline 
HOG2-dep \cite{refref39}        & \xmark  & \cmark   & \xmark  &  59.83 \\    
HOG2-dep+pose \cite{refref39}   & \xmark  & \cmark   & \cmark  &  66.78 \\ 
HON4D \cite{refref40}               & \xmark  & \cmark   & \xmark  &  70.61 \\ 
Novel View \cite{refref41}          & \xmark  & \cmark   & \xmark  &  69.21  \\ 
\hline
1-layer LSTM \cite{Zhua2016}        & \xmark  & \xmark   & \cmark  &  78.73 \\
2-layer LSTM \cite{Zhua2016}        & \xmark  & \xmark   & \cmark  &  80.14 \\ 
\hline 
Moving Pose \cite{refref59}         & \xmark  & \xmark   & \cmark  &  56.34 \\ 
Lie Group \cite{Vemulapalli2014}    & \xmark  & \xmark   & \cmark  &  82.69 \\ 
HBRNN \cite{Du2015}                & \xmark  & \xmark   & \cmark  &  77.40 \\ 
Gram Matrix \cite{refref61}         & \xmark  & \xmark   & \cmark  &  85.39 \\ 
TF    \cite{refref11}               & \xmark  & \xmark   & \cmark  &  80.69 \\  
\hline 
JOULE-color \cite{refref18}         & \cmark  & \xmark   & \xmark  &  66.78 \\ 
JOULE-depth \cite{refref18}         & \xmark  & \cmark   & \xmark  &  60.17 \\ 
JOULE-pose \cite{refref18}         & \xmark  & \xmark   & \cmark  &  74.60 \\ 
JOULE-all \cite{refref18}           & \cmark  & \cmark   & \cmark  &  78.78 \\ 
\hline 
Huang et al. \cite{Huangcc2017}     & \xmark  & \xmark   & \cmark  &  84.35 \\ 
Huang et al. \cite{ref23}           & \xmark  & \xmark   & \cmark  &  77.57 \\  
\hline 
HAN  \cite{Liu2021}   & \xmark  & \xmark   & \cmark & 85.74 \\
  \hline
  \hline
Our baseline GCN                   & \xmark  & \xmark   & \cmark  & 88.17                                                  
\end{tabular}}
\end{center}
\caption{Comparison of our baseline GCN (not label-efficient) against related work on the FPHA database.}\label{compare2}
\end{minipage}
\end{table}
\noindent {\bf Implementation details \& baseline GCNs.} All GCNs have been trained using the Adam optimizer for $2700$ epochs. The batch size is $200$ for SBU and $600$ for FPHA.  A momentum of $0.9$ is used, and the global learning rate $\nu$  is dynamically adjusted based on the loss Eq.~\ref{eqqqqq}'s rate of change.  Specifically, $\nu$ is decreased by a factor of $0.99$ when the rate of loss change {\it increased}, and increased by a factor of $1/0.99$ {\it otherwise}.  Training is performed on a GeForce GTX 1070 GPU with $8$ GB memory.  No dropout or data augmentation techniques are employed. For SBU, the architecture of our GCN comprises three ``mono-head attentions + (8 filters) convolutions'' layers followed by one fully connected and a classification layer. The  GCN architecture for FPHA is relatively heavier (for a GCN), and differs from SBU in the number of convolutional filters (16 filters instead of 8). Both architectures, on the SBU and the FPHA benchmarks, are accurate  (see Tables.~\ref{tab222}-\ref{compare2}), and our goal is to make them label-efficient while being {\it as close as possible} to their initial accuracy. \\
\noindent{\bf Performances, comparison \& ablation.} Tables~\ref{table21}-\ref{table22}  show a comparison and an ablation study of our method both on the SBU and the FPHA datasets. According to the observed results, when our display model is run on the ambient space, the accuracy is relatively high, and sometimes overtakes comparative display selections by a noticeable margin. When using the latent space, we observe a  further gain of our method. This clearly shows the impact of our model and its extra gain when combined with the latent space. Extra comparison of our method against other display selection strategies also shows a substantial gain. Indeed, our method is compared against different strategies  used  as display selection (instead of our proposed display model),  namely random, diversity \cite{zhang2022multi} and uncertainty~\cite{zhao2023uncertainty}, all with our GCN learning. From the observed results in tables~\ref{table21}-\ref{table22}, the impact of our method  is significant for different settings  and for  equivalent labeling rates. We also observe that random is already performant (as widely known, see for instance~\cite{Burr2009} and references therein) mainly when the sample size is relatively large (45\%). In contrast, with relatively smaller sizes (15\%), random is less performant so  more principled selection strategies are required.

Note that random and diversity are not capable of sufficiently refining classifications, whereas uncertainly allows us to refine classifications but without enough diversity. Besides, all these comparative methods suffer at some extent from the rigidity of the selected displays (which are taken from a fixed pool). Our display model, in contrast, allows us to learn flexible exemplars, constrained in the latent space of the proposed invertible and stable GCNs, with a positive impact on performances including at frugal labeling regimes. 

\begin{table}[h]
 \begin{center}
\resizebox{0.79\columnwidth}{!}{
  \begin{tabular}{cll}    
   \rotatebox{0}{Labeling rates}  &     \rotatebox{0}{Accuracy (\%)}  & \rotatebox{0}{Observation}  \\
 \hline
  \hline
    100\%    &    98.40     & Baseline GCN (not label-efficient)\\
     \hline
    \multirow{5}{*}{\rotatebox{0}{45\%}}     & \underline{89.23}   & wo display model (random display)  \\
                                             &  \underline{89.23}   & + display model + ambient space (our) \\
                                             &   \bf93.84   & + display model + latent space (our) \\
                                             &67.69   & uncertainty (margin-based) \\                                                                                                                                                                                   & 83.07  & diversity (coreset-based)\\                                        
    \hline
      \multirow{5}{*}{\rotatebox{0}{30\%}}   & 80.00  & wo display model (random display)  \\
                                             & \underline{86.15}  & + display model + ambient space (our) \\
                                             & \bf87.69   & + display model + latent space (our) \\
                                             & 61.53  & uncertainty (margin-based) \\                                                                                                                                                                              & 83.07   & diversity (coreset-based)\\                                        
    \hline 
        \multirow{5}{*}{\rotatebox{0}{15\%}} & \underline{69.23}  & wo display model (random display) \\
                                             & \bf75.38  & + display model + ambient space  (our) \\
                                             & \bf75.38  & + display model + latent space (our) \\
                                             & 56.92  & uncertainty (margin-based) \\                                                                                                                                                                              & 66.15  & diversity (coreset-based) \\                                        
    \hline
  \end{tabular}}
\end{center}
\caption{This table shows detailed performances and ablation study on SBU for different  labeling rates. Here ``wo'' stands for ``without''. Best results are shown in bold and second best results underlined.}\label{table21}
\end{table}
 \begin{table}[h]
 \begin{center}
\resizebox{0.79\columnwidth}{!}{
  \begin{tabular}{cll}    
   \rotatebox{0}{Labeling rates}  &     \rotatebox{0}{Accuracy (\%)}  & \rotatebox{0}{Observation}  \\
 \hline
  \hline
    100\%    &    88.17     & Baseline GCN (not label-efficient)\\
     \hline
    \multirow{3}{*}{\rotatebox{0}{45\%}}     & \underline{75.47} & wo display model (random display)  \\
                                             & 72.52  & + display model + ambient space (our)  \\
                                             & \bf75.65 & + display model + latent space (our) \\
                                             & 63.30   & uncertainty (margin-based) \\                                                                                                                                                                            & 70.26   & diversity (coreset-based)\\         
                                  
    \hline
      \multirow{3}{*}{\rotatebox{0}{30\%}}   & \bf67.47   & wo display model (random display)  \\
                                             & 61.21   & + display model + ambient space (our) \\
                                             & \underline{63.65}   & + display model + latent space (our) \\
                                             & 56.17   & uncertainty (margin-based) \\                                                                                                                                                                             & 62.08    & diversity (coreset-based)\\                                        
    \hline 
        \multirow{3}{*}{\rotatebox{0}{15\%}} &  40.52   & wo display model (random display)  \\
                                             &  45.21        & + display model + ambient space (our)  \\
                                             & \bf49.21    & + display model + latent space (our) \\
                                             & 41.73    & uncertainty (margin-based) \\                                                                                                                                                                             & \underline{46.26}   & diversity (coreset-based) \\                                       
      \hline 
  \end{tabular}}
\end{center}
\caption{Same caption as table~3, but for FPHA.}\label{table22}
\end{table}

\section{Conclusion}
We introduce in this paper a label-efficient method for skeleton-based action recognition built upon graph convolutional networks (GCNs).  The strength of our contribution resides in the design of a new acquisition function as the optimum of an objective function mixing representativity, diversity and uncertainty. We further enhance this design by making our GCNs stable and invertible thereby transforming input data into latent and more readily learnable spaces. The efficacy and superior performance of our proposed method are demonstrated through extensive experiments on two challenging skeleton-based recognition datasets.
\section*{Appendix} 
\begin{proof}[\bf Sketch of the Proof (Proposition 2)]  \it 
Given a metric space $(A, d_A)$, where $d_A$ denotes the metric on the set $A$ (by default  $d_A$ is taken as $\ell_2$ and $A$ as $\mathbb{R}^p$); considering a subsumed version of our GCNs, and using the Lipschitz continuity, one may write
\begin{eqnarray*}
d_A(f(\phi_1^{1}),f(\phi_2^{1})) & =&  d_A(g_L(\W_L^\top \phi_1^{L-1}), g_L(\W_L^\top \phi_2^{L-1}))  \\
    & \leq  &  u  \ d_A(\W_L^\top \phi_1^{L-1},\W_L^\top \phi_2^{L-1})  \\
    & \leq  & u .  \|\W_L\|_A  \ \ d_A(\phi_1^{L-1},\phi_2^{L-1}) \\
    & \leq & u^{L-1} \|\W_L\|_A \dots   \|\W_2\|_A \ \  d_A(\phi_1^{1},\phi_2^{1}),
\end{eqnarray*}
being $\phi_1^{1}$, $\phi_2^{1}$ two network inputs. As  $\{\W_\ell\}_\ell$ are orthonormal, it follows that $\|\W_\ell\|_A =1$, \\ and $d_A(f(\phi_1^{1}),f(\phi_2^{1})) \leq K \ d_A(\phi_1^{1},\phi_2^{1})$ with $K=u^{L-1}$.\\ 
Similarly for  $f^{-1}$, given an output $\phi^{L}$, we have $f^{-1}(\phi^{L})=\phi^{1}$ with  $\phi^{\ell-1}=(\W_\ell^\top)^{-1} g_\ell^{-1}(\phi^{\ell})$. Hence, considering two network outputs $\phi_1^{L}$,  $\phi_2^{L}$ one may write
\begin{eqnarray*}
 d_A(f^{-1}(\phi_1^{L}),f^{-1}(\phi_2^{L}))=&  d_A((\W_2^\top)^{-1} g_2^{-1}(\phi_1^{2}), (\W_2^\top)^{-1} g_2^{-1}(\phi_2^{2}))  \\
       \leq & \|(\W_2^\top)^{-1}\|_A \ \  d_A( g_2^{-1}(\phi_1^{2}), g_2^{-1}(\phi_2^{2}))  \\
     \leq  &  \|(\W_2^\top)^{-1}\|_A \ (1/l)  \ d_A(\phi_1^{2},\phi_2^{2}) \\
     \leq & \prod_{\ell} \|(\W_\ell^\top)^{-1}\|_A \ (1/l)^{L-1} \  d_A(\phi_1^{L},\phi_2^{L}).
\end{eqnarray*} 
As $\{\W_\ell\}_\ell$ are orthonormal, it follows that $\|(\W_\ell^\top)^{-1}\|_A =1$, and $d_A(f^{-1}(\phi_1^{L}),f^{-1}(\phi_2^{L})) \leq M d_A(\phi_1^{L},\phi_2^{L})$ with  $M = (1/l)^{L-1}$ 
\end{proof}


\begin{thebibliography}{1}
\bibitem{aaaaa9}  H. Sahbi.  A particular Gaussian mixture model for clustering and its application to image retrieval.   Soft Computing 12 (7), 667-676
\end{thebibliography}
\end{document}